\documentclass[conference]{IEEEtran}
\IEEEoverridecommandlockouts
\usepackage[sort]{cite}
\usepackage{amsmath,amssymb,amsfonts}
\usepackage{algorithmic}
\usepackage{graphicx}
\usepackage{multirow}
\usepackage{tablefootnote}
\usepackage[hidelinks]{hyperref}
\usepackage{textcomp}
\usepackage{changes}
\usepackage{balance}
\usepackage{xcolor}
\usepackage{amsmath}
\def\BibTeX{{\rm B\kern-.05em{\sc i\kern-.025em b}\kern-.08em
T\kern-.1667em\lower.7ex\hbox{E}\kern-.125emX}}

\usepackage{enumitem}
\setlist[itemize]{leftmargin=10pt}

\usepackage{threeparttable}
\usepackage{hyperref}

\begin{document}

\title{Automatic Speech Recognition with BERT and CTC Transformers: A review}

\makeatletter 
\newcommand{\linebreakand}{%
  \end{@IEEEauthorhalign}
  \hfill\mbox{}\par
  \mbox{}\hfill\begin{@IEEEauthorhalign}
}
\makeatother 

\author{\IEEEauthorblockN{Noussaiba Djeffal}
\IEEEauthorblockA{\textit{Speech and Signal Processing Lab.} \\
\textit{USTHB University}\\
Algiers, Algeria\\
ndjeffal@usthb.dz}
\and
\IEEEauthorblockN{Hamza Kheddar}
\IEEEauthorblockA{\textit{LSEA Lab., Faculty of Technology} \\
\textit{ University of MEDEA}\\
Medea 26000, Algeria \\
kheddar.hamza@univ-medea.dz}
\and 

\IEEEauthorblockN{Djamel Addou}
\IEEEauthorblockA{\textit{Speech and Signal Processing Lab.} \\
\textit{USTHB University}\\
Algiers, Algeria\\
daddou@usthb.dz }

\linebreakand
\and 
\IEEEauthorblockN{Ahmed Cherif Mazari}
\IEEEauthorblockA{\textit{LSEA Lab, Faculty of Science} \\
\textit{ University of MEDEA}\\
Medea 26000, Algeria \\
mazari.ahmedcherif@univ-medea.dz}

\and 
\IEEEauthorblockN{Yassine Himeur}
\IEEEauthorblockA{\textit{College of Engineering and Information } \\
\textit{Technology, University of Dubai}\\
Dubai, UAE \\
yhimeur@ud.ac.ae}
}

\makeatletter

\def\ps@headings{%
\def\@oddhead{\parbox[t][\height][t]{\textwidth}{\flushleft

\noindent\makebox[\linewidth]
}
\vspace{0.5cm}
\hfil\hbox{}}%
\def\@oddfoot{\MYfooter}%
\def\@evenfoot{\MYfooter}}

\def\ps@IEEEtitlepagestyle{%
\def\@oddhead{\parbox[t][\height][t]{\textwidth}{\centering
2023 2nd International Conference on Electronics, Energy and Measurement (IC2EM 2023)\\

}\hfil\hbox{}}%

\def\@oddfoot{ 979-8-3503-1424-3/23/\$31.00 \textcopyright 2023 IEEE \hfil 
\leftmark\mbox{}}%
\def\@evenfoot{\MYfooter}}

\maketitle

\begin{abstract}
This review paper provides a comprehensive analysis of recent advances in automatic speech recognition (ASR) with bidirectional encoder representations from transformers BERT and connectionist temporal classification (CTC) transformers. The paper first introduces the fundamental concepts of ASR and discusses the challenges associated with it. It then explains the architecture of BERT and CTC transformers and their potential applications in ASR. The paper reviews several studies that have used these models for speech recognition tasks and discusses the results obtained. Additionally, the paper highlights the limitations of these models and outlines potential areas for further research. All in all, this review provides valuable insights for researchers and practitioners who are interested in ASR with BERT and CTC transformers.
\end{abstract}


\section{Introduction}
Traditional methods of speech recognition rely on maximum a posteriori probability estimation, which involves transforming the acoustic speech characteristics into word sequences through four steps: feature extraction, acoustic modeling, language modeling, and word sequence decoding. Feature extraction involves essential data extraction from the input signal using algorithms such as Mel-frequency cepstral coefficients (MFCC) \cite{kheddar2019pitch} and perceptual line spectral pairs (LSP) \cite{kheddar2022high}. The acoustic modeling stage utilizes deep neural networks and hidden Markov models to map the acoustic frame to the phonetic state at each input time, optimized for the phonetic classification error per frame. Language modeling is designed to model the most probable sequences of words, regardless of acoustics \cite{2023towards}. The use of transformers in speech recognition involves several steps, such as : (i) preprocessing of the audio signal, to extract essential features like log Mel filterbank energies, (ii) an acoustic model based on a self-attention mechanism to model the temporal relationships between acoustic features, and (iii) a language model trained on a large amount of text data to capture long-term dependencies between words. The language model takes a sequence of words as input and predicts the probability distribution of the next word in the sequence. Finally, the decoding process involves finding the most likely word sequence, given the output of both the acoustic and language models.

To sum up, transformers such as BERT, and connectionist temporal classification (CTC) based ASR is a recent advancement in speech recognition that uses the self-attention mechanism to simultaneously perform feature extraction, acoustic modeling, language modeling, and decoding in a single network. On the other hand, the transformer architecture is a neural network model that is designed to process sequential data by attending to relevant context information. Transformers have demonstrated promising outcomes in ASR and are expected to play a crucial role in future advancements in this field. 



\subsection{Related work and our contribution}
 In\cite{2021BERTT } the authors explore various methodologies for detecting emotions in text using BERT and its variants. They thoroughly outline their approaches, contributions, achieved accuracies, and also discuss the limitations or weaknesses of their models. However, our review focuses on ASR rather than emotion using BERT. While the authors primarily mention BERT base, BERT large, RoBERTa, DistillBERT, and cross-lingual language model (XLM), which are the BERT variants employed in their research, we expand on their contributions by incorporating additional models such as ALBERT, and ELECTRA.  
 The work in \cite{osf2020} focuses on providing a tutorial and survey on the attention mechanism, transformers, BERT, and GPT. It explains various concepts such as the attention mechanism, transformers, and their components. However, our review focuses on both BERT and CTC transformer applications, specifically in ASR. In \cite{2022ACM} the survey discusses the impressive performance of transformer models like BERT, GPT, RoBERTa, and T5 across various language tasks such as text classification, machine translation, and question answering. Additionally, the article also explores their applications in computer vision. However, our review focuses on BERT and CTC transformers within the domain of ASR, distinguishing our research from other domains discussed in the aforementioned survey. In \cite{2023surevy} encompasses various speech-related domains such as ASR, speech synthesis, speech translation, speech para-linguistics, speech enhancement, and other applications. The authors of the survey identify and discuss the challenges that transformers face in these domains. In our review, similar to the survey, we address these challenges, but we extend the scope by including additional transformers like ELECTRA, ALBERT, and CTC Transformers specifically in the context of ASR.

 \subsection{Paper structure}
The rest of the paper is organized as follows: Section \ref{sec2} presents the methodology used to create this review. Section \ref{sec3} provides an overview of the preliminaries on CTC and BERT. Section \ref{sec4} defines BERT and categorizes articles based on the utilization of BERT in ASR, highlighting the advancements made in this area. Section \ref{sec5} defines CTC and classifies articles based on the application of CTC in ASR, showcasing the progress achieved in this domain. Section \ref{sec6} discusses potential future directions. Finally, the paper concludes in section \ref{sec7}.

\section{Review Methodology and Analysis}
\label{sec2}
This review focuses mainly on two distinct categories of papers: The first category examines BERT-based ASR systems, the second category of papers explores CTC-based ASR techniques. The initial search was done using related keywords for transformers in ASR, namely: BERT and ASR, CTC and ASR, or BERT and CTC and ASR. 
 A search was performed on scientific databases indexed at least in Scopus and available in IEEE Xplore and Springer, science direct, and others. Besides, arXiv papers are taken into consideration that have many citation and impact, which are known for their extensive coverage and relevance to ASR. Additionally, Google Scholar was utilized to include a broader range of publications, including gray literature, which can provide valuable insights for a systematic review. Only the most widely used methods and implementations were included to ensure modularity. The focus is on papers that reported new and unique applications within specific domains, avoiding repetition. Emphasis is given to papers published in high-quality journals with a significant impact factor. The search has been conducted until 2023 to gather the most recent available information at the time of the review.

\section{Preliminaries}
\label{sec3} 

\subsection{Dataset}
Several datasets have been utilized in various ASR tasks in the existing literature. Table \ref{tab01} provides a compilation of some of these datasets that have been used specifically for BERT and CTC-based ASR applications, along with their specific characteristics \cite{2023towards}.
\begin{table*}[t!]
\centering
\caption{a list of publicly available datasets commonly used in transformers BERT and CTC-based ASR research: }
\label{tab01}
\begin{tabular}{p{2cm}p{1.2cm}p{14cm}} \hline Dataset & Used by & Description \\ [0.6mm]
\hline
FGC & \cite{luo2020spoken,kuo2020audio}& Is a collection of data specifically focused on the task of spoken multiple-choice question answering in Mandarin Chinese. This dataset was created for the formosa grand challenge (FGC) competition held in 2018.
\\[0.6mm]
AMI database &\cite{chiu2021cross}  & Is a widely used and publicly available dataset in the field of multimodal interaction research. Researchers use the augmented multiparty interaction (AMI) database for tasks such as speech recognition, speaker diarization, language understanding, dialogue systems, and multimodal analysis. \\[0.6mm]
CNNDM, How2, TED & \cite{kano2021attention} & The CNNDM dataset is a large-scale dataset primarily used for text summarization tasks. It consists of news articles collected from the websites of cable news networks, and daily mail, the How2 dataset is designed for the task of instructional video captioning and translation. Specifically, it provides textual descriptions of "how-to" videos, which are instructional videos demonstrating various tasks and activities. The TED corpus is a collection of summarized TED talks; the corpus was constructed by linking TED talks from the TEDLIUM corpus with their corresponding summaries.\\[0.6mm]
DSTC2 & \cite{ganesan2021n} & The dialog state tracking challenge 2 (DSTC2) dataset is a widely used benchmark dataset in the field of dialog systems and spoken language understanding.  \\[0.6mm]
IWSLT2011 & \cite{chen2021discriminative,chen2020controllable} & Serves as a benchmark for the task of spoken language translation, dataset contains approximately 25 hours of recorded speech. This duration includes speech data in multiple languages, such as English, French, Spanish, and German.\\ 
 FSC, ATIS &\cite{jiang2021knowledge} & The frame-semantic corpus (FSC) and airline travel information system (ATIS) datasets are both widely used in the field of natural language processing (NLP) and dialogue systems. The total duration of the FSC dataset is approximately 130 hours. 
 \\[0.6mm]
Fisher-CallHome Spanish & \cite{inaguma2021fast} & 
It is a combination of two data corpora: Fisher and CallHome. The Fisher corpus consists of multilingual telephone recordings and was originally collected for research in ASR. The CallHome corpus is a dataset of telephone conversations in different languages, it consists of approximately 300 hours of recorded speech.\\[0.6mm]
 LibriSpeech &\cite{fan2021improved} & It consists of speech data from audiobooks available in the LibriVox project. The dataset comprises approximately 1,000 hours of audio recordings, with a sampling rate of 16 kHz. It is a collection of high-quality speech data obtained by sampling audiobooks in the project.\\ [0.6mm]
Aishell1 &\cite{song2021non,fan2021improved,nozaki2021relaxing} & It comprises a large collection of Mandarin Chinese speech recordings from multiple speakers. It contains approximately 178 hours of speech data from around 400 speakers, covering a wide range of accents, ages, and genders.\\ [0.6mm]
WSJ & \cite{futami2020distilling,higuchi2020mask,nozaki2021relaxing,moritz2020all} & Typically refers to the Wall street journal dataset, which is a commonly used benchmark dataset in NLP and information retrieval research, contained around 80 hours of transcribed speech.\\[0.6mm]
CSJ  &\cite{fujita2020toward,futami2020distilling} & Typically refers to the corpus of spontaneous Japanese. The CSJ dataset contained approximately 570 hours of recorded speech.\\
TEDLIUM2 &\cite{fujita2020toward,nozaki2021relaxing}& Is a widely used benchmark dataset in the field of ASR, created by the spoken language systems (SLS) group at the University of Cambridge. \\[0.6mm]
DCASE & \cite{moritz2020all} & It includes binaural recordings captured in 15 different sound environments or settings. These settings represent distinct acoustic scenes and cover a variety of audio environments. \\
\hline     

\end{tabular}
\end{table*}

\subsection{Metrics}
The ASR research community has employed several methods to evaluate the quality and generalizability of ASR techniques. Besides the famous metrics that are commonly employed in ML and DL, such as accuracy, F1-score, recall, and precision \cite{kheddar2023deepSteg}, other metrics are used including word error rate (WER) and character error rate (CER), and real-time factor (RTF), are thoroughly described in\cite{2023towards}.

\section{BERT-based ASR}
\label{sec4}
BERT, developed by Devlin et al. in 2019 \cite{devlin2018bert}, is a pre-training model for NLP tasks that utilizes transformer encoders \cite{kheddar2024transformers}. It consists of two phases: pre-training for language understanding and fine-tuning. These latter are for specific tasks like sentiment analysis, question answering, text summarization, and more. During pre-training, BERT employs masked language modeling (MLM) and next-sentence prediction (NSP). MLM involves masking some words in sentences and reconstructing them using the surrounding context during training. NSP helps BERT understand the relationship between two sentences by predicting if the second sentence follows the first. BERT was trained on 16GB of text data from the books corpus datasets and English Wikipedia. After pre-training, the model is fine-tuned for a specific task by replacing BERT's output layers. This fine-tuning process is faster since only the model parameters, excluding the output parameters, are learned from scratch. There are two versions of BERT: BERT-base and BERT-large. BERT-base consists of 12 transformer encoder blocks with 12-head self-attention layers and 768 hidden layers, resulting in approximately 110 million parameters. BERT-large has 24 transformer encoder blocks with 24-head self-attention layers and around 340 million parameters. BERT-large achieves higher accuracies but requires more computational resources compared to BERT-base \cite{2021BERTT}. However, BERT has a few notable limitations. Firstly, it is primarily designed for monolingual classifications, meaning its optimal performance is achieved when working with a single language. While it is possible to fine-tune BERT for multilingual tasks, its effectiveness may be somewhat diminished compared to its performance on monolingual tasks. Secondly, the length of input sentences can also present challenges. BERT has a maximum token limit, typically set at 512 tokens, which means longer sentences need to be truncated or split into smaller segments, potentially losing some contextual information \cite{mazari2023bert}.  Table \ref{tab:2} presents a summary of the performance achieved in the cited papers compared to other systems,  including the metrics used to evaluate the results, and the source code availability. Figure \ref{fig1} illustrates BERT variants featuring transformers along with the attention layers they are built upon.

\subsection{ BERT-based SMCQA framework}
The authors in \cite{luo2020spoken} developed a framework called MA-BERT for spoken multiple-choice question answering (SMCQA) task, which uses a combination of multi-turn audio-extractor hierarchical CNNs (MA-HCNNs) and BERT to extract acoustic-level and text-level information, respectively, from speech data. The proposed framework outperformed various state-of-the-art systems. However, the scheme in \cite{kuo2020audio} proposes a novel audio-enriched BERT-based (aeBERT) framework for improving performance on the SMCQA task, where syllables, questions, and choices are all given in speech. Besides, the method proposes incorporating acoustic-level information from the speech input to enhance the accuracy of SMCQA systems.  The resulting audio-enriched BERT-based SMCQA framework shown to outperform various state-of-the-art systems by a large margin.

\begin{figure}[bt!]
\centering
\includegraphics [scale=0.3]{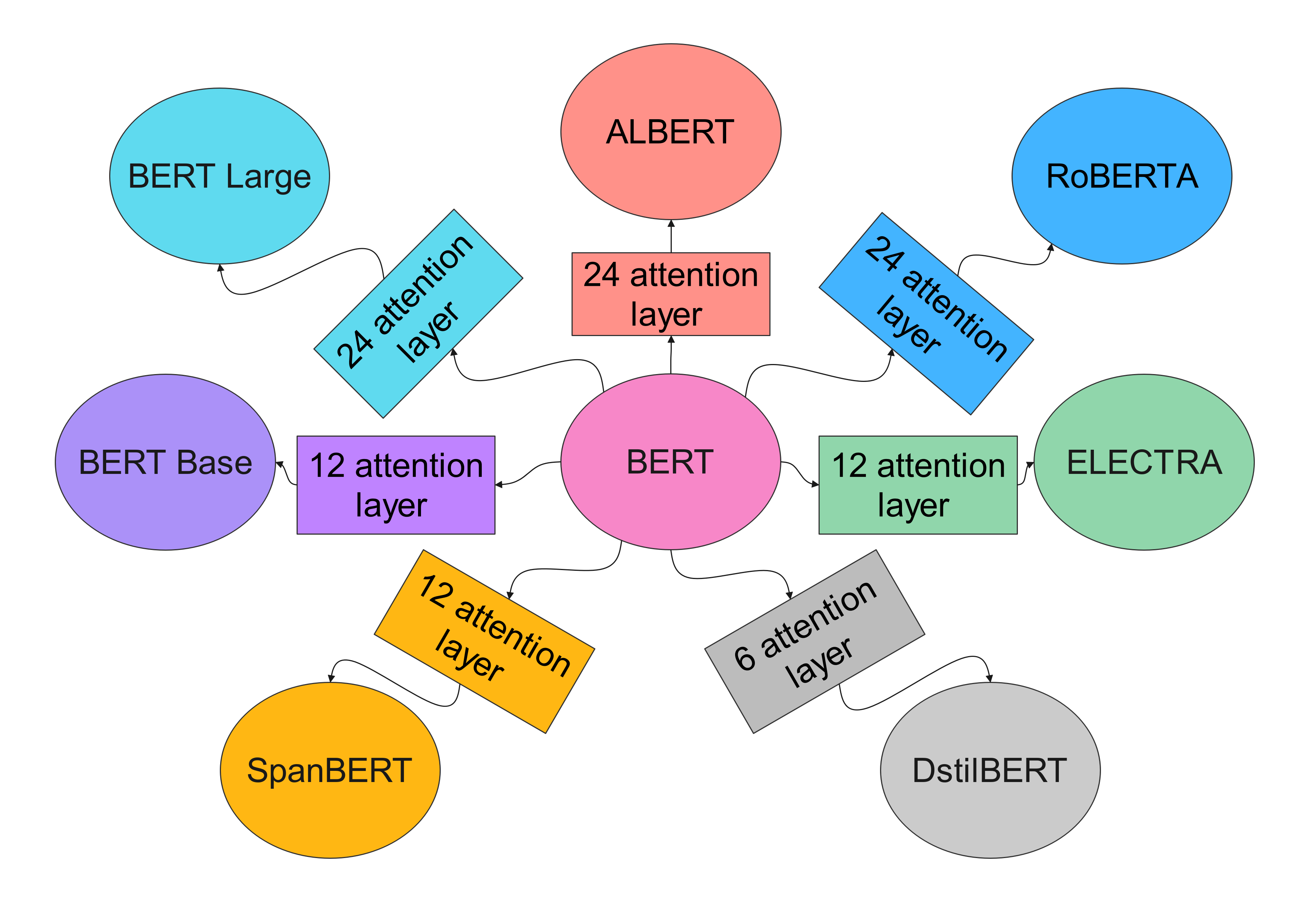}
\caption{ Types of BERT with transformer encoder layers.}
\label{fig1}
\end{figure}

\subsection{BERT-based reranking framework }
 Chiu et al.\cite{chiu2021cross} propose a BERT n-best reranking framework that incorporates cross-utterance information signals using a graph convolutional network (GCN) to model historical utterances for better ASR performance. The approach addresses the limitations of recurrent neural network (RNN) and LSTM-based language models (LMs) in capturing complex global structural dependencies among utterances. Nevertheless, the study in \cite{chiu2021innovative} introduces a new implementation of BERT-based contextualized language models specifically for reranking the n-best hypotheses generated by ASR systems. The approach frames the n-best hypothesis reranking as a prediction problem, aiming to predict the oracle hypothesis with the lowest WER.

\subsection{ BERT-based model for speech summarization}
 Kano et al. \cite{kano2021attention} suggest a novel text summarization (TS) method that combines sub-word embedding vectors and posterior values from an ASR system. They incorporate an attention-based fusion module into a pre-trained BERT module for improved summarization. This fusion module aligns and merges multiple ASR hypotheses. The researchers then perform experiments on speech summarization using both the How2 and TED dataset. In \cite{weng2021effective} The authors of the paper enhance a BERT-based model for speech summarization in three ways: incorporating confidence scores into sentence representations to address ASR errors, augmenting sentence embeddings with additional features, and validating the model's effectiveness on a benchmark dataset compared to classic summarization methods. The goal is to improve the model's performance and overcome challenges caused by imperfect ASR.
\subsection{BERT-based model for distilling the knowledge}
Futami et al. \cite{futami2020distilling}  propose a method to improve ASR using a combination of a (sequence-to-sequence) seq2seq model and BERT as an external language model. The seq2seq model is enhanced with both left and right context through knowledge distillation from BERT which generates soft labels to guide the training. Additionally, context beyond the current utterance is leveraged as input to BERT. The proposed method is evaluated on the CSJ, showing significant improvements in ASR performance compared to the seq2seq baseline. This method surpasses alternative approaches in LM applications like n-best rescoring and shallow fusion with not requiring any additional inference cost. Jiang, B et al. \cite{jiang2021knowledge} suggest a method for end-to-end intent classification using speech, which does not rely on an intermediate ASR module. It leverages the transformer distillation method to transfer knowledge from a transformer-based language model BERT to a transformer-based speech model for intent classification. A multi-level transformer-based teacher-student model is designed, and knowledge distillation is performed across attention and hidden sub-layers of different transformer layers. The proposed method achieves a high level of accuracy in intent classification and showcases superior performance and resilience in acoustically degraded conditions when compared to the baseline method.

\subsection{BERT, RoBERTa, XLM-RoBERTa, and ELECTRA models }

Ganesan et al. \cite{ganesan2021n} propose a method to improve the performance of spoken language understanding (SLU) systems by using concatenated n-best ASR alternatives as input to transformer models, such as  BERT XLM-RoBERTa on DSTC2 dataset \cite{henderson2014second}.
In their paper, Chen et al. \cite{chen2021discriminative} introduce a discriminative self-training method that incorporates weighted loss and discriminative label smoothing for improving punctuation prediction in ASR output transcripts, the authors utilize extensive unlabeled spoken language data, which lacks punctuation, such as transcripts employed for training ASR systems. They employ self-training techniques to enhance robust baseline models built on BERT, RoBERTa, and ELECTRA.

\subsection{HuBERT and LightHuBERT models}
In their study \cite{ren2022speech}, the authors introduce a novel speech pre-training method called "HuBERT-AP." This approach utilizes patterns derived from target codes as the training signal to facilitate the model in acquiring improved acoustic features. The patterns, referred to as "acoustic pieces," are constructed based on the sentence piece outcomes of the original HuBERT target codes, and are highly relevant to phonemized natural language, making them beneficial for audio-to-text tasks. The proposed method is evaluated on the LibriSpeech ASR task, and is shown to be significantly more effective than previous strong baselines. However, the authors in \cite{wang2022lighthubert} propose LightHuBERT, a compressed version of the HuBERT model, which is a self-supervised speech representation learning model. LightHuBERT is designed as a once-for-all transformer compression framework. To automatically discover desired architectures through pruning structured parameters, the researchers create a transformer-based supernet that encompasses numerous weight-sharing subnets. They also employ a two-stage distillation strategy to leverage contextualized latent representations from HuBERT. Experimental results on ASR and the SUPERB benchmark demonstrate that LightHuBERT surpasses HuBERT in ASR tasks while reducing parameters by 29\%. Furthermore, LightHuBERT achieves a compression ratio of 3.5 times in three SUPERB tasks, albeit with a slight loss in accuracy.
 \subsection{NorBERT and Speech-BERT models}
Rugayan et al. \cite{rugayan2022semantically} propose a robust evaluation metric, aligned semantic distance (ASD), for Norwegian ASR systems. They leverage semantic information modeled by a transformer-based LM and employ dynamic programming techniques to measure the similarity between reference and hypothesis text. ASD utilizes NorBERT embeddings to compute the optimal alignment and obtain the minimum global distance. This distance is then normalized by the length of the reference embedding vector. Additionally, the researchers present results using another metric called semantic distance (SemDist), and they compare the performance of ASD with SemDist. The authors in \cite{fan2019neural} introduced a neural model called speech-BERT, which combines a bidirectional transformer LM with a neural zero-inflated beta regression approach. This approach is specifically designed to be conditioned on speech features.  To fine-tune speech-BERT, the authors utilized a pre-training strategy known as token-level masked language modeling. Additionally, they incorporated a zero-inflated layer into the model to effectively handle the mixture of discrete and continuous outputs.

\subsection{BERT-based language models}

Chang et al. \cite{chang2021context} introduce an innovative network called the context-aware transformer transducer (CATT), which enhances the performance of transformer-based ASR systems by leveraging contextual signals. The authors propose a context-biasing network based on multi-head attention, which is trained alongside other sub-networks of the ASR system. Various techniques are explored to encode contextual data and generate the ultimate attention context vectors. To encode the contextual information and facilitate network training, both BLSTM and pre-trained BERT models are utilized. The researchers in \cite{fohr2021bert} propose two deep neural network (DNN) models to improve ASR by modeling long-term semantic relations. They employed as input features to their DNN model two things: (i) dynamic contextual embeddings are derived from BERT, a transformer-based model specifically designed for acoustic tasks. (ii) Additionally, linguistic features are incorporated into the system. Moving forward, the scheme proposed in \cite{priya2022multilingual} discusses the linguistic diversity in India and the need for speech recognition in regional languages. The paper suggests the creation of an advanced ASR system based on deep sequence modeling, aiming to address the challenges posed by low-resource languages. The proposed model incorporates an enhanced spell corrector component. The performance of the proposed system is assessed using metrics such as WER and sequence match ratio. Notably, the experimental results demonstrate promising outcomes, with an average WER of 0.62. The latter result proves the importance of spell correction in ASR systems and the use of a transformer-based LM for performance improvement.

\section{CTC-based ASR}
\label{sec5}
CTC is a variant of the transformer architecture that is used in seq-2seq learning tasks, particularly in ASR. The CTC transformer combines the concepts of the CTC loss function and the transformer architecture, which are both powerful tools for sequence modeling. The CTC loss function is commonly used in ASR to align the predicted sequence with the ground truth sequence by taking into account the presence of blank symbols and repeated characters. In the following, a brief summary of the proposed approaches-based CTC transformer. Table \ref{tab:2} presents a summary of the performance achieved in the cited papers compared to others systems,  including the metrics used to evaluate the results, and the source code availability. Figure  \ref{fig2} shows a CTC variation, elucidating its intended purpose and objectives.
 \begin{figure}[htbp]
\centering
\includegraphics [scale=0.3]{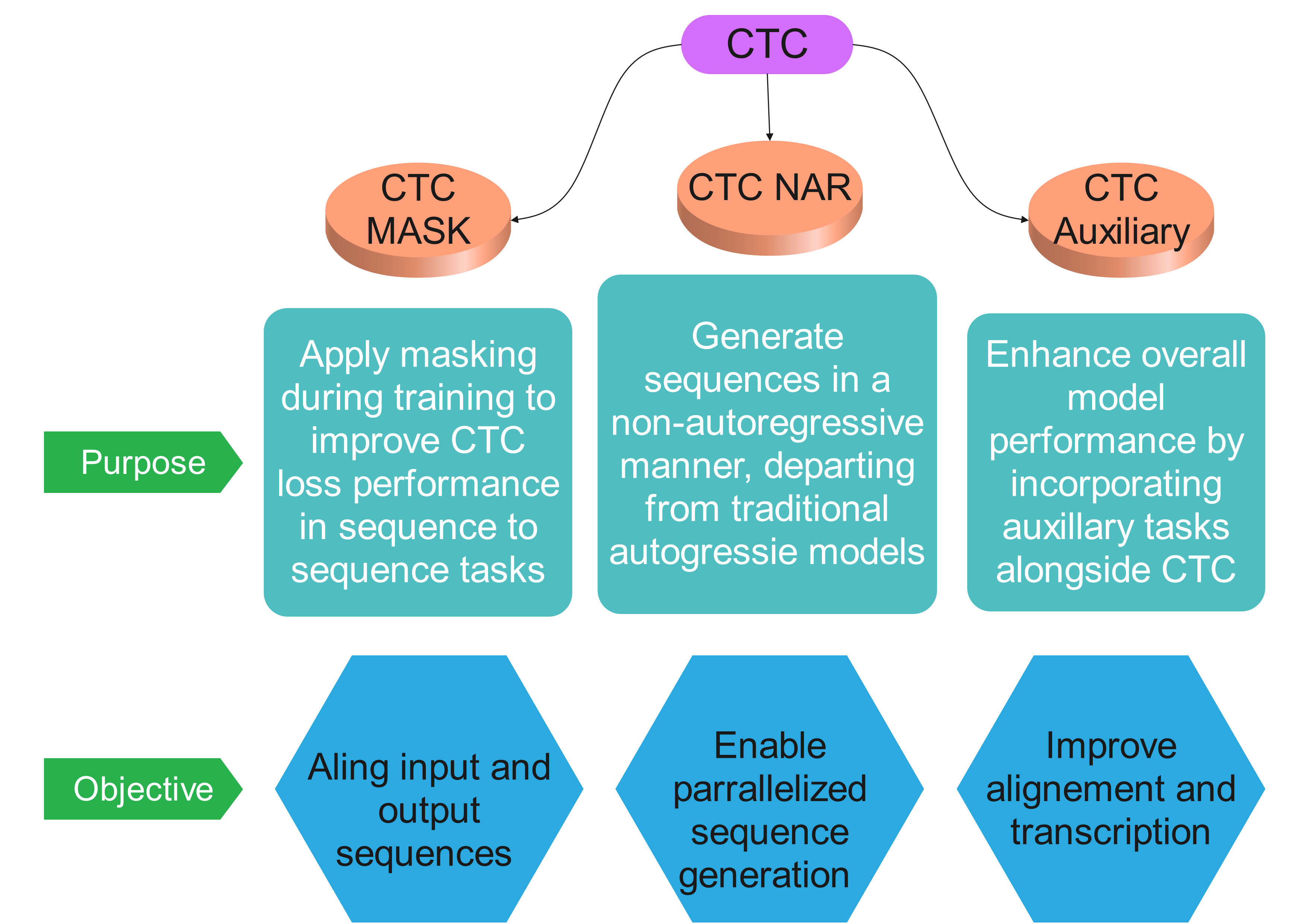}
\caption{ Purpose and objective  of CTC.}
\label{fig2}
\end{figure}

\subsection{Mask CTC}

The proposed method, detailed in \cite{higuchi2020mask}, consists of a novel non-autoregressive end-to-end ASR called mask CTC. This framework generates a sequence by refining the outputs of the CTC model, which is a popular method used for ASR, while autoregressive models generate one token at a time and require as many iterations as the output length. Non-autoregressive models offer the advantage of generating tokens simultaneously in a fixed number of iterations, resulting in substantial reductions in inference time. The mask CTC model employs a training methodology that combines a transformer encoder-decoder architecture with simultaneous training of mask prediction and CTC during inference. Initially, the target sequence is initialized with the greedy CTC outputs. Afterward, tokens with low confidence are selectively masked using the CTC predictions. By taking into account the conditional interdependence among output tokens, the model predicts the masked low-confidence tokens using the high-confidence tokens.

\subsection{NAR CTC}
Inaguma et al in \cite{inaguma2021fast} propose a faster version of the multi-decoder (MD) end-to-end speech translation model called Fast-MD. The MD model decomposes the overall speech translation task into ASR and machine translation sub-tasks, but its decoding speed is not fast enough for real-world applications. Fast-MD generates hidden intermediates (HI) by NAR decoding based on CTC outputs followed by an ASR decoder. The scheme employs sampling CTC outputs during training to reduce a mismatch in the ASR decoder. The authors also suggest that adopting the conformer encoder and intermediate CTC loss can further boost the model's quality without sacrificing decoding speed.
Song et al. \cite{song2021non} propose a solution to the accuracy degradation problem faced by NAR transformer models in ASR. The proposed solution is a CTC-enhanced NAR transformer that refines the predictions of the CTC module to generate the target sequence.
The paper \cite{fan2021improved} presents improvements to the end-to-end CTC alignment-based single-step non-autoregressive transformer (CASS-NAT) for speech recognition. The proposed methods include applying convolution augmented self-attention blocks to the encoder and decoder modules, expanding the trigger mask for each token to increase CTC alignment robustness, and using iterated loss functions to enhance gradient updates. Fujita et al. in \cite{fujita2020toward} proposed a method for non-autoregressive  ASR streaming input or long recording. They used an insertion-based model that jointly trained  CTC and achieved better accuracy with fewer iterations using transformer with greedy decoding. The authors suggested combining audio segmentation and non-autoregressive ASR into a single neural network. This integration leverages the CTC component of the insertion-based model, utilizing causal self-attention implemented through block self-attention, similar to the transformer XL. Experimental outcomes demonstrated that the proposed approach achieved a favorable trade-off between accuracy and RTF when compared to both the autoregressive transformer and CTC baseline models.

\subsection{Auxiliary CTC and End-to-end CTC}

The method introduced by the authors in \cite{nozaki2021relaxing} offers a means to enhance CTC-based ASR models by loosening the assumption of conditional independence in CTC. The method involves training a CTC-based ASR model with auxiliary CTC losses in intermediate layers. Predictions from these layers are accumulated and conditioned on in subsequent layers, resulting in improved performance compared to a standard CTC model across multiple ASR corpora. Furthermore, the proposed method achieves comparable performance to a strong auto-regressive model with beam search on the TEDLIUM2 corpus and the AISHELL-1 corpus, while being at least 30 times faster in decoding speed. Andrusenko et al. in \cite{andrusenko2020exploration} explore different end-to-end ASR systems for the largest open-source Russian language data set – OpenSTT. They compare existing end-to-end approaches, including joint CTC/Attention, RNN-transducer, and transformer, with a strong hybrid ASR system based on the so-called LF-MMI and TDNN-F acoustic model. Is the performance of each system is evaluated on three available validation sets, including phone calls, YouTube, and books. The paper \cite{gelin2021simulating} adapted an end-to-end transformer acoustic model to the speech of children learning to read, with the aim of enhancing ASR performance for this challenging task. They used transfer learning with a small amount of child speech and multi-objective training with a CTC function. They also proposed a method of data augmentation for reading mistakes, where they simulated word-level repetitions and substitutions with phonetically or graphically close words. The authors analyzed the performance of their model and showed that both the CTC multi-objective training and the data augmentation with synthetic repetitions assisted the attention mechanisms better identify children's disfluencies.

\subsection{CTC-Based Other Approaches }

In their work \cite{chen2020controllable}, Chen et al. introduce the controllable time-delay transformer (CT-Transformer) model, which addresses both punctuation prediction and disfluency detection tasks in real-time. These tasks are crucial for enhancing transcript readability and enabling subsequent applications. The CT-Transformer model incorporates a mechanism for selectively freezing partial outputs with adjustable time delays to meet the real-time constraints imposed by downstream applications. Experimental results demonstrate that the proposed approach surpasses previous state-of-the-art models in terms of F-scores, while also achieving competitive inference speed on benchmark datasets such as IWSLT2011 \cite{che2016punctuation} and an in-house Chinese annotated dataset.  Moritz et al. in \cite{moritz2020all} describe the development and implementation of an "all-in-one" (AIO) acoustic model based on the transformer architecture. The AIO model is designed to simultaneously solve the problems of ASR audio tagging (AT), and acoustic event detection (AED), using shared parameters across all tasks. The authors argue that this approach more closely mimics the way the human auditory system processes sound signals from different sources. The integration of the transformer model with CTC enables the enforcement of monotonic ordering and the utilization of timing information for both ASR and AED tasks. The AIO transformer model consistently outperforms all baseline systems in recent DCASE challenge tasks, showcasing its aptness for comprehensive transcription of acoustic scenes, encompassing speech recognition and identification of other acoustic events. Xiao et al. in \cite{xiao2021automatic} propose a new framework for an automatic voice query service AVQS to improve the accuracy of response for multi-accented Mandarin users. The problem addressed is that many dialect areas in China make it necessary for the AVQS to respond to users with a single acoustic model in ASR, limiting its accuracy. The proposed framework uses a fusion feature comprising i-vector and filter-bank acoustic features to train a transformer-CTC, which is then used to construct an end-to-end ASR. Additionally, a keyword-matching algorithm based on fuzzy mathematics theory is proposed to further enhance the accuracy of the response.

\begin{table*}[h!]
\caption{List of the surveyed state-of-the-art studies with their advantages and disadvantages.}
\label{tab:2}
\begin{tabular}
{llp{1.5cm}p{1.5cm}lp{0.5cm}p{10cm}}
\hline 
Ref. & Year & Category & Compared to & Metric& Code \newline avail.? & Result / Improvement obtained / Comments / Advantages and/or disadvantages \\[0.6mm]
\hline
\cite{luo2020spoken} &2020 & MA-BERT & BERT-RNN &  accuracy  & No & 80.34\%, improvement of 2.5\%, the proposed MA-BERT It is an ideal framework for leveraging both acoustic-level and text-level features in the SMCQA  task.\\[0.6mm]
\cite{chiu2021cross} & 2021 & HPBERT(10)+ GCN(10) & HPBERT(10) & WER & No & 16.13\%, reduction over 0.14\%, the global information captured by the GCN enhances the representation of historical utterances, leading to improved reranking performance.\\

\cite{ganesan2021n} & 2021 & n-Best-ASR BERT& WCN-BERT STC & F1-scores &  No & 87.80 \%, improvement of 1.6\%, the N-Best ASR Transformer offers improved performance over baselines, excels in low data regimes, and provides accessibility to users of third-party ASR APIs.\\

\cite{chen2021discriminative} & 2021 & RoBERTa-wwm-base+Disc-ST & RoBERTa-wwm-base  & F1-scores & Yes\footnotemark[3] & 60.2\%, Discriminative Self-Training improves F1 from 59.6 to 60.2 (+0.6), this approach significant improvement on punctuation prediction over strong baselines including  RoBERTa models.\\

\cite{jiang2021knowledge}  & 2021 & STD-BERT & Baseline-2 & accuracy & Yes\footnotemark[2] & 99.10\%, improvement of 0.23\%, the experimental results show improved accuracy after incorporation of transformer-based knowledge distillation.\\

 \cite{fan2021improved} & 2021 & Improved CASS-NAT & Conformer AT & RTF &  No & 0.018, RTF degradation (from 0.081 to 0.018), This suggests that the enhanced CTC alignment-based CASS-NAT achieves comparable performance to AT.\\

\cite{song2021non} & 2021&NAR-Transformer& AR-Transformer  & RTF & Yes\footnotemark[4]  & 0.0037, results show a Non-autoregressive Trans- former with CTC-enhanced decoder achieve 50x faster decoding speed than a strong AR baselin.\\

\cite{higuchi2020mask} & 2020& Mask CTC & Non-autoregressive
CTC & CER & Yes\footnotemark[5]  & 4.96\%, a reduction over 0.53\%, the experimental comparisons demonstrated that Mask CTC outperformed the standard CTC model while maintaining the decoding speed fast.\\

\cite{moritz2020all} & 2020 &  AIO Transformer & Baseline system & F1-scores & No & 51.4\%, experiments demonstrate that the AIO Transformer achieves better performance compared to all baseline systems of various re- cent DCASE challenge tasks.\\

\cite{fujita2020toward} & 2020 & KERMIT-Integrated CTC & ART-Integrated CTC & RTF & No & 0.38, RTF degradation (from 1.15 to 0.38), The results indicate that the method successfully achieved a reasonable balance between RTF and performance when compared to the baseline autoregressive Transformer and connectionist temporal classification approaches.\\

\cite{chiu2021innovative} & 2021 & TPBERT & PBERT & WER &  No & 20.49\%, reduction over 0.78\%, the advantages of TPBERT lie in its effective use of BERT-based ASR N-best hypothesis. \\

\cite{wang2022lighthubert} & 2022 &LightHuBERT & DistilHuBERT & PER & Yes\footnotemark[1]  & 4.71\%, reduction over 11.56\%. This demonstrates the superior performance of LightHuBERT compared to DistilHuBERT.\\

\cite{fohr2021bert} & 2021 &  BERT$_{\text{alsem}}$ and GPT-2 scores & Baseline system & WER & No & 34.4\%, reduction of 2.7\%, the optimal outcomes are attained by combining rescoring techniques that utilize BERT and GPT-2 scores.\\

\cite{andrusenko2020exploration} & 2020 & TDNN-F-LF-MMI & CTC-Attention & WER & Yes\footnotemark[5]  & 33.5\%, reduction over 5.4\%, The hybrid model continues to outperform end-to-end systems in terms of performance on phone call validation. By incorporating an external NNLM for hypotheses rescoring within the hybrid system, a reduction in WER is achieved across all validation sets.\\

\cite{gelin2021simulating} & 2021 & Transformer +CTC +Sub+Rep  & Transformer (baseline)& PER &No & 19.90\%, a reduction over 3\%, A comprehensive analysis demonstrates that both the multi-objective training with CTC and the augmentation using synthetic repetitions effectively enhance the ability of attention mechanisms to detect disfluencies in children's speech.\\

\cite{xiao2021automatic} & 2021 &Transformer-CTC &BLSTM-CTC &SER &  No & 65.05\%, reduction of 13.1\%, the proposed framework can effectively improve the whole response accuracy of AVQS for heavy accented Mandarin speech.\\

\hline
\end{tabular}
\end{table*}
\footnotetext[1]{\href{https://github.com/ mechanicalsea/lighthubert}{https://github.com/lighthubert} ( Accessed: July 2023)}

\footnotetext[2]{\href{ https://github.com/kaituoxu/Speech- Transformer/blob/master/src/transformer/loss.py}{https://github.com/kaituoxu/Speech-Transformer} (Accessed: July 2023)}

\footnotetext[3]{\href{ https://github.com/google-research/bert }{https://github.com/google-research/bert} (Accessed: July 2023)} 

\footnotetext[4]{\href{https://github.com/espnet/espnet}{https://github.com/espnet/espnet} (Accessed: July 2023)}

\footnotetext[5]{\href{ http://www.dev.voxforge.org/projects/Russian/export/2500/Trunk/AcousticModels/etc/msu_ru_nsh.dic}{http://www.dev.voxforge.org/AcousticModels} (Accessed: July 2023)}

\footnotetext[6]{\href{https://github.com/espnet/espnet}{https://github.com/espnet/espnet} (Accessed: July 2023)}

\section{Future directions}
\label{sec6}

\subsection{ BERT and ChatGPT}
\begin{itemize}
    \item \textbf{Improved contextual coherence:} By combining ChatGPT ability to generate human-like responses with BERT strong contextual understanding, the integration enhances the coherence and relevance of the chat responses by leveraging BERT knowledge of bidirectional dependencies in text.

\item \textbf{Enhanced language comprehension:} BERT extensive pre-training on a large corpus enables it to understand language nuances effectively. When integrated with ChatGPT, it improves the model's language understanding capabilities, enabling it to comprehend user inputs, handle complex queries, and provide more accurate and context-aware responses.
\item \textbf{Effective handling of ambiguity and multiple meanings:} ChatGPT can sometimes struggle with phrases that have multiple interpretations. By incorporating BERT contextual representation, which considers the surrounding context, the integrated model becomes better at disambiguating such phrases and generating responses that are more accurate and appropriate in context.

\end{itemize}

\subsection{CTC and ChatGPT}
\begin{itemize}

    \item \textbf{Enhanced language generation:} By integrating ChatGPT into CTC, the speech output generated becomes more natural and engaging due to ChatGPT  proficiency in generating human-like responses.
\item \textbf{Context-aware speech Generation:} Incorporating ChatGPT ability to understand contextual cues into CTC enables the model to generate speech that is more coherent and relevant by considering the surrounding context.

\item \textbf{Versatile text-to-speech applications:} CTC is widely used in text-to-speech systems. Integrating ChatGPT with CTC expands the capabilities of TTS applications, making them more flexible and adaptable. This allows for interactive and dynamic speech generation by leveraging ChatGPT conversational capabilities.

\item \textbf{Enhanced personalization and user interaction:} ChatGPT excels in personalized conversations, and when combined with CTC, it enables the integrated model to generate speech that adapts to user preferences. This results in more interactive and engaging interactions, leading to a highly personalized user experience.
 \end{itemize}

\section{Conclusion}
\label{sec7}

Transformers play a crucial role in ASR by capturing contextual information and long-range dependencies. They improve accuracy by considering the entire context and utilizing attention mechanisms to focus on relevant information. Pre-trained models like BERT, RoBERTa, and ELECTRA have proven effective in transfer learning for ASR, benefiting from knowledge acquired on large-scale datasets. Additionally, the CTC method enables end-to-end training, handles variable-length inputs, incorporates label-smoothing regularization, integrates with language models, and supports online and streaming ASR applications. CTC is a flexible and effective approach for transcribing speech signals, contributing to robust and accurate ASR systems applied to diverse domains, such as biomedical \cite{essaid2024artificial}.

In this survey, the idea of incorporating ChatGPT into the BERT and CTC frameworks is proposed, opening new avenues for research and development. By leveraging ChatGPT's conversational abilities and natural language understanding, it is suggested to enhance BERT and CTC capabilities. This integration aims to improve ASR performance, accuracy, and contextual understanding, leading to advanced speech recognition applications.
\balance
\bibliographystyle{IEEEtran}

\bibliography{references.bib}

\end{document}